\documentclass[runningheads]{llncs}
\usepackage[T1]{fontenc}

\usepackage{graphicx}
\usepackage{listings}
\usepackage{bbold}
\usepackage{colortbl}
\usepackage{xcolor}
\usepackage{comment}
\usepackage{xifthen}
\usepackage{xspace}
\usepackage{hyperref}

\begin{document}

\title{Agent-Exploitation Affordances: From Basic to Complex Representation Patterns}
\titlerunning{Agent-exploitation affordances: representation patterns}
%
%

\author{Bastien Dussard\inst{1}\orcidID{0009-0001-8800-8853} \and
Aurélie Clodic\inst{1}\orcidID{0009-0009-6484-8143} \and Guillaume Sarthou\inst{1}\orcidID{0000-0002-4438-2763}}
\authorrunning{B. Dussard et al.}

\institute{
LAAS-CNRS, Universit\'e de Toulouse, CNRS, Toulouse, France \\ 
\email{firstname.surname@laas.fr}
}

\renewcommand\lstlistingname{Description}
\renewcommand\lstlistlistingname{Descriptions}
\renewcommand{\thelstlisting}{\arabic{lstlisting}}

\lstdefinestyle{customDescription2}{
  language=C,
  stringstyle=\itshape\color{green!25!black},
  keywordstyle=\bfseries,
  morekeywords={rdf, rdfs, type, domain, subPropertyOf, range, hasSubtask, DecompositionUsedBy, subClassOf, hasDecomposition, Cut_hasParameter, A, V, K}
}

\maketitle              
\begin{abstract}

In robotics, the capability of an artificial agent to represent the range of its action possibilities, i.e. affordances, is crucial to understand how it can act on its environment. While functional affordances, which refer to the use of tools and objects, have been broadly studied in knowledge representation, the implications of a social context and the presence of other agents have remained unexplored in this field. Consequently, in the field of social robotics, a multi-agent context enables the agents to engage in new actions that are potentially complementary to their individual capabilities, leading to the perspective of agent-exploitation. This work focuses on the concept of cooperative affordance within the realm of social affordances. Cooperative affordances refer to situations where agents interact with each other to extend their action possibilities range. From this definition, this paper proposes a tractable ontological representation of this concept with the aim of making it usable by an artificial agent. Expanding on those elementary patterns, we illustrate the effectiveness of these representations by combining them to depict a diverse range of scenarios.

\end{abstract}
\section{Introduction}

For an artificial agent to act efficiently upon its environment, it is essential that it possesses knowledge regarding its interaction capabilities with non-agentive entities, which enables it to bring about changes in its surroundings. Indeed, since for a robot to successfully complete its task requires that it understands how to bring the environment to a desired different state, knowledge over the action possibilities it affords in a given environment is key.
 
In the literature, this concept is commonly referred to as \textbf{affordance} and was first introduced by Gibson in~\cite{gibson2014}. Over the years, this concept has been refined in multiple fields given its open-to-interpretation nature, such as ecological psychology~\cite{stoffregen2003affordances}, industrial design~\cite{norman2002design}, or robotics~\cite{csahin2007afford}. As each of those formalisms leveraged the concept for different purposes, they introduced refinements given their usage such as “disposition”, “perceivability”, or “effect” which captured important aspects. Nevertheless, two major views co-exist for representing affordances, either as a concept (Turvey in~\cite{turvey1992ecological}) or as a relation between concepts (Chemero in~\cite{chemero2003outline}). Those views differ given the author's perspective, either environmental or agent/observer~\cite{csahin2007afford}.


In this work, we build upon the formalism introduced by Chemero in~\cite{chemero2003outline}, which outlines the relational nature of affordances. However, one could see that given the purpose of our work, it contradicts with the anti-representationalist perspective of most work in ecological psychology~\cite{chemero2007gibsonian}. Nevertheless, as discussed in~\cite{zech2017computational}, \textit{“roboticists generally extract features as a basis for affordance detection and learning, thereby implicitly building an internal representation.”}, motivating that the computational view over affordances is meaningful in robotics. Both Chemero's original formalism and its extension for computational models consider affordances as emerging relations between agent's capabilities and entities' dispositions.
 
While a capability refers to \textit{“the ability to carry out a type of activity”}~\cite{solano2016ontology}, a disposition can be referred to as \textit{“the property of a thing that is a potential”}~\cite{turvey1992ecological}. Those two concepts work in pairs which enable action possibility. For instance, an agent capable of grasping and an object holding a graspable disposition can be matched together, resulting in the emergence of an affordance relation.

Although some affordances can stem from a single entity's disposition (e.g. graspable), some entities can require interaction with other non-agentive entities for their dispositions to be actualized. For example, a lock needs a key to actualize its openable disposition. This mutual need for interaction was referred to as \textbf{reciprocal dispositions} by Martin in~\cite{martin2010mind}. It states that complementarity can occur between dispositions, and consequently can \textit{“partner for a mutual manifestation that is their common product”}.
Functional affordances represent relations linking an agent to an entity given its dispositions or multiple entities given their reciprocal dispositions. However, the affordance concept extends beyond those latter and encompasses a wider range of domains. Indeed, while the functional affordances' nature is intrapersonal, meaning they involve only a single agent, a social setting enables interpersonal action possibilities emerging from the presence of multiple agents. Those latter can be referred to as \textbf{social affordances} and were defined as \textit{“possibilities for social interaction or possibilities for action that are shaped by social practices and norms”} by Carvalho in~\cite{carvalho2020social}. It highlights two influence factors: culture (cultural affordance~\cite{ramstead2016cultural}) and social conventions (normative affordance~\cite{heras2016affordances}). It also outlines that the involvement of multiple agents can impact the action range, thus the affordance relations. Indeed, a multi-agent context enables the agents to engage in new actions that are potentially complementary to their individual capabilities.

The mutual involvement of agents can enable new actions that are useful to complete tasks in goal-directed behavior, and thus the possibility of \textbf{agent-exploitation} emerges through interpersonal affordances. In the context of goal-directed interpersonal affordances, it is possible to distinguish between different categories of affordances. The upper-level category of \textbf{cooperative affordances} encompasses opportunities for action between multiple agents toward a goal with potentially independent actions, which do not necessarily have to impact the environment directly (e.g. an agent communicating with another one to help it prepare a recipe). Moving down a level, the concept of collaborative affordance, as outlined in ~\cite{bardram2018collaborative}, emerges when there are opportunities for agents to act toward a goal, even though their actions may differ, but both engage directly in the action (e.g. two agents working on a piece of assembly). Joint affordances~\cite{davis2010perceiving} are even more precise in that they refer to opportunities for action that arise toward a shared goal but with similar actions and on the same entity (e.g. agents lifting a heavy table together).
We introduce the former, cooperative affordance, as “affordance relation enabled by agents that can interact, thereby extending the range of action possibilities they afford”. From this definition, one can see that collaborative or joint affordances are included since in both cases, the agents act together toward a common goal. Nevertheless, joint affordances differ as they involve shared opportunities for simultaneous action, while collaborative affordances involve coordinated and cooperative interactions between agents to achieve a common goal. Consequently, joint affordances require a temporal synchronization between agents, which we chose not to include in our representation since this particular case requires more precise factors to be properly represented.



The main contribution of this paper is an applicable \textbf{pattern to represent cooperative affordances} which aims at highlighting how affordances can emerge through the involvement of other agents. This contribution is strengthened by the possibility of combining elementary patterns in order to represent diverse situations in a tractable manner.


In Sec.~\ref{sec:related_work}, we briefly discuss related work. We provide in Sec.~\ref{sec:elem} a representation of functional affordances and dispositional match with regard to Chemero's formalism and finally, we introduce a representation of cooperative affordances. In Sec.~\ref{sec:cooperative}, we build upon those elementary representations to show how their combinations can represent various situations. In Sec.~\ref{sec:combined}, we provide an overview of how such patterns enable to represent several pathways for the actualization a given disposition. Finally, we conclude in Sec.~\ref{sec:conclusion} by discussing possible future work.

\section{Related Work}
\label{sec:related_work}

In robotics, the concept of affordance has been used from many different perspectives over the years~\cite{ijcai2021p590}, but few approaches tackled the representation of affordances in robotic knowledge bases. However, the representation of the affordance concept is essential as it conveys the idea of action possibilities. Yet, most work focused on affordances between an agent and objects (functional affordance), but to the best of our knowledge, the action possibilities provided by having several agents in a given environment (social affordance) have not been tackled before.

In ontologies, the concept of functional affordance is often represented implicitly, without conceptualizing the core idea of conditions enabling the affordances. Those implicit representations range from an inheritance over classes~\cite{OROlemaignan2017artificial} (e.g. \textit{Mug, isA,  CanBeManipulated}), a relation between an agent and an object~\cite{OROSUgonccalves2015knowledge} (e.g. \textit{bob, canGrab, mug}), or a relation between an object and an affordance~\cite{su2017improving} (e.g. \textit{mug, hasAffordance, pourability}).

Nevertheless, some work aimed at conceptualizing this idea, which was mostly tackled with the \textbf{functional affordance}~\cite{hartson2003cognitive} perspective.
This concept's core idea is that some entities can provide a functionality by their usage, which results in a change of the environment. Varadarajan et al. in~\cite{varadarajan2011ontological} implemented this concept by a binding between geometrical features and functional affordances. Thanks to \textbf{Conceptual Equivalence Classes}, they represented object concepts given their part's geometry and their related affordances. For example, a knife is represented as composed of a sharp-edged entity with an incision-ability affordance (blade) and a flat entity with a grasp-ability affordance (handle).
While this functional affordance representation involves only a single entity, other affordances can involve several entities such as a pen being used with a paper sheet. This combination of multiple entities has been referred to as \textbf{affordance dualities} in~\cite{varadarajan2012afrob} and highlights the reciprocality between specific aspects of complementary entities. Given the pen/paper sheet example, the reciprocality stems from their respective engrave-ability and display-ability affordances. Since both of those entities' features relate to non-agentive entities, we rather refer to them as dispositions instead of affordances, and the complementing pair of dispositions as reciprocal dispositions~\cite{martin2010mind}. 

Such a reciprocal perspective over affordances/dispositions has been tackled by Beßler et al in~\cite{bessler2020formal}, building upon the \textbf{bearer/trigger} roles. The bearer's role represents the entity holding the disposition to be actualized by a suitable other entity, the trigger. For instance, in a cleaning task, a dirty plate would be the bearer and a dishwasher would be the trigger. While this representation conveys the reciprocal perspective, it does not leverage the agentive aspect. Indeed, this formalism aims at answering competency questions such as \textit{“What can this be used with?”} rather than being instantiated in a given environment. In~\cite{toyoshima2018formal}, Toyoshima and Barton introduced an ontological formal characterization of affordances that leverages the agent capabilities and the entities' dispositions. The proposed pattern builds upon the reciprocal nature of those concepts and allows for a more generic representation of functional affordances. Although this formalization provides the frame for precise representation, its structure is not suitable to represent affordances involving more entities. Indeed, the pattern is designed for one-to-one matching and requires an important number of individuals dedicated to each affordance.

To sum up, most work tackling the representation of affordances follow Turvey's perspective and focus on functional affordances. To the best of our knowledge, none of the existing proposals account for the interpersonal affordances that emerge from the presence of multiple agents in a given environment. Those latter enable multi-agent goal-directed behavior through agent-exploitation, hence cooperative affordances.

\section{Elementary Affordances}
\label{sec:elem}

\begin{figure*}[!t]
    \centering
    \includegraphics[width=\linewidth]{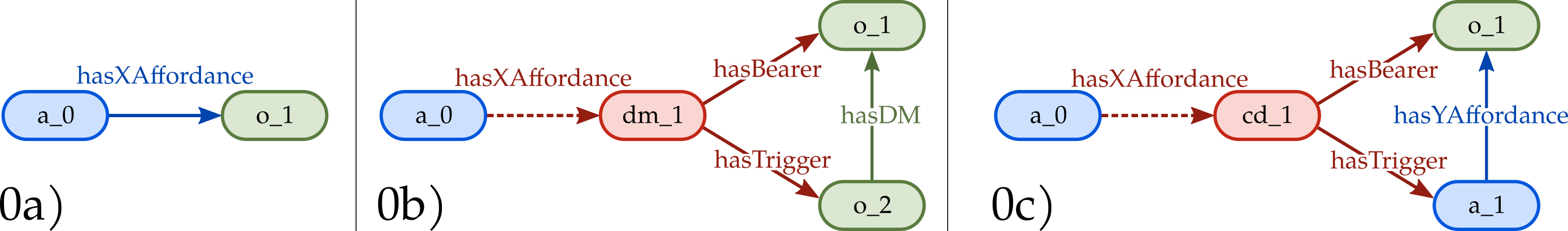}
    \caption{Three ontological patterns to represent: 0a) functional affordance, 0b) functional affordance toward dispositional match, 0c) cooperative affordance. The affordance properties X and Y are generic and can be different. Agents' capabilities and objects' dispositions are omitted for readability.}
    \label{fig:basic_patterns}
\end{figure*}

\subsection{Agent-Object Affordance}

Given Chemero's formalism, an affordance can be seen as a relation between an agent's capabilities and an entity's dispositions. A direct representation of such a view is depicted in Fig.~\ref{fig:basic_patterns}.0a with a relation between the agent and the entity, since an affordance conveys the idea of an agent's action possibility toward an entity \textit{(a\_0, hasXAffordance, o\_1)}. The \textit{X} could refer to any affordance and is meant as a generic representation of the relational concept.

An example of such an affordance from an agent toward an object could be the physical affordance of grasping, if the agent has the required capabilities to grasp this particular entity regarding its dispositions. If the conditions are satisfied, then the ontological relation \textit{(agent, hasGraspingAffordance, mug)} would be created.
The representation of the used capabilities is voluntarily omitted in this work as previous work has already presented possible representation, as Dussard et al. in~\cite{dussard2023ontological}. In this work, the authors describe capabilities as enabled by the agent's components. In a similar manner, as the capabilities of an agent stem from the set of components it owns, dispositions of an object can stem from its sub-parts~\cite{sarathy2016cognitive}.

\subsection{Agent-Dispositional Match Affordance}

While sub-parts of a single entity can provide an object with different dispositions, each of them actualizable by an agent's capabilities, different entities can have dispositions of interacting with each other. Those latter do share a reciprocal nature, meaning that they can be paired together as a sum of matching dispositions to bring a change in the environment. Taking inspiration from Toyoshima and Barton in~\cite{toyoshima2018formal}, we can consider this matching as a sum of dispositions and represent it through the use of relation reification. Unlike the authors, we chose to introduce only a single individual which we refer to as \textbf{Dispositional Match} (DM), representing the compositional disposition created by the match. As illustrated in Fig.~\ref{fig:basic_patterns}.0b, the agent no longer has an affordance toward the individual entities but rather toward the Dispositional Match, keeping track of which dispositions of the entities were necessary to create the affordance relation. Considering the terminology used by Be{\ss}ler et al in~\cite{bessler2020formal}, we refer to the matched entities as the bearer and the trigger of the DM since each of the involved entities either has the role of actualizing the disposition or to be actualized.

An example of such a pattern could be illustrated by a scenario in which there is an agent capable of grasping and motion planning, a knife having the \textit{Cutting} disposition, and a tomato having the \textit{Cuttable} disposition. From this scenario, one can see that the dispositions of the knife and the tomato are reciprocal. Thus, the agent capable of acting upon both entities has the affordance of cutting the tomato with the knife. As the affordance stems from both the entities' dispositions, the affordance \textit{(agent, hasCuttingAffordance, dm\_knife\_tomato)} emerges.

\subsection{Affordance Agent-Cooperative Dispositional Match}

Since in a DM, each entity has a role either to be actualized or to actualize given its dispositions (bearer or trigger), one could view an affordance relation between an agent and an entity in a similar way. Indeed, thanks to communication and thus interaction, an agent can rely on another agent to act on the environment. Rather than having a direct affordance toward an entity, it can have the affordance to cooperate with another agent to query this latter to act. We refer to such social affordance involving other agents to act on entities as a \textbf{cooperative affordance}. An ontological representation is depicted in Fig.~\ref{fig:basic_patterns}.0c.
Similarly to a DM stemming from two entities interacting together given their reciprocal dispositions, one could consider an agent and an entity linked by an affordance relation as a complementary pair providing new action possibilities. Using a similar representation, we no longer consider a Dispositional Match entity but rather a \textbf{Cooperative Disposition} (CD) entity. Indeed, while the DM individual represents the sum of complementary dispositions, the cooperative disposition aims to represent the sum of the object's dispositions and the interaction dispositions of the involved agent. An agent having the right set of interaction capabilities could thus have a cooperative affordance with a CD. 

To illustrate, let us consider a scenario in which an agent needs to lift a heavy dishwasher but lacks the necessary capabilities. If another agent is present who can effectively lift this object (lifting affordance), then this agent and the dishwasher can be considered a complementary pair that can be interacted with to bring about the desired change. The representation of such a pair may be referred to as \textit{cd\_agent\_1\_dishwasher} and it encompasses the object's actualizable disposition (\textit{Liftable}) as well as the agent's interactional dispositions (\textit{VerbalCommunication}). Consequently, if the first agent is capable of engaging in the relevant interactions, it gives rise to the emergence of the affordance relation (\textit{agent\_0, hasLiftingAffordance, cd\_agent\_1\_dishwasher}).

\subsection{Example of a Practical Implementation}

These patterns can be represented either in OWL, Turtle, or Prolog. An example of such instantiation in Turtle is provided in~\ref{list:turtle_example}. It illustrates the cooperative affordance emerging between a robot and a human given the functional affordance of cutting a tomato. Through interaction with the human, which has the required capabilities, the pr2 robot has the affordance of cutting the tomato, even though it might not have had the affordance to do it by itself.

\begin{lstlisting}[float, floatplacement=H, frame=single, basicstyle=\scriptsize\ttfamily, caption={\label{list:turtle_example} Description of a cooperative affordance relation between agents (robot and human) to cut down a tomato.}, captionpos=b, style=customDescription2]
<#human>
    a :Agent ;
    :hasCuttingAffordance <#tomato> .
<#cd_human_k>
    a :CooperativeDispositionalMatch ;
    :hasTrigger <#human> ;
    :hasBearer <#tomato> .
<#pr2>
    a :Agent ;
    :hasCuttingAffordance <#cd_human_k> .
\end{lstlisting}

\section{Complex Cooperative Affordances}
\label{sec:cooperative}

\begin{figure*}[]
    \centering
    \includegraphics[width=\linewidth]{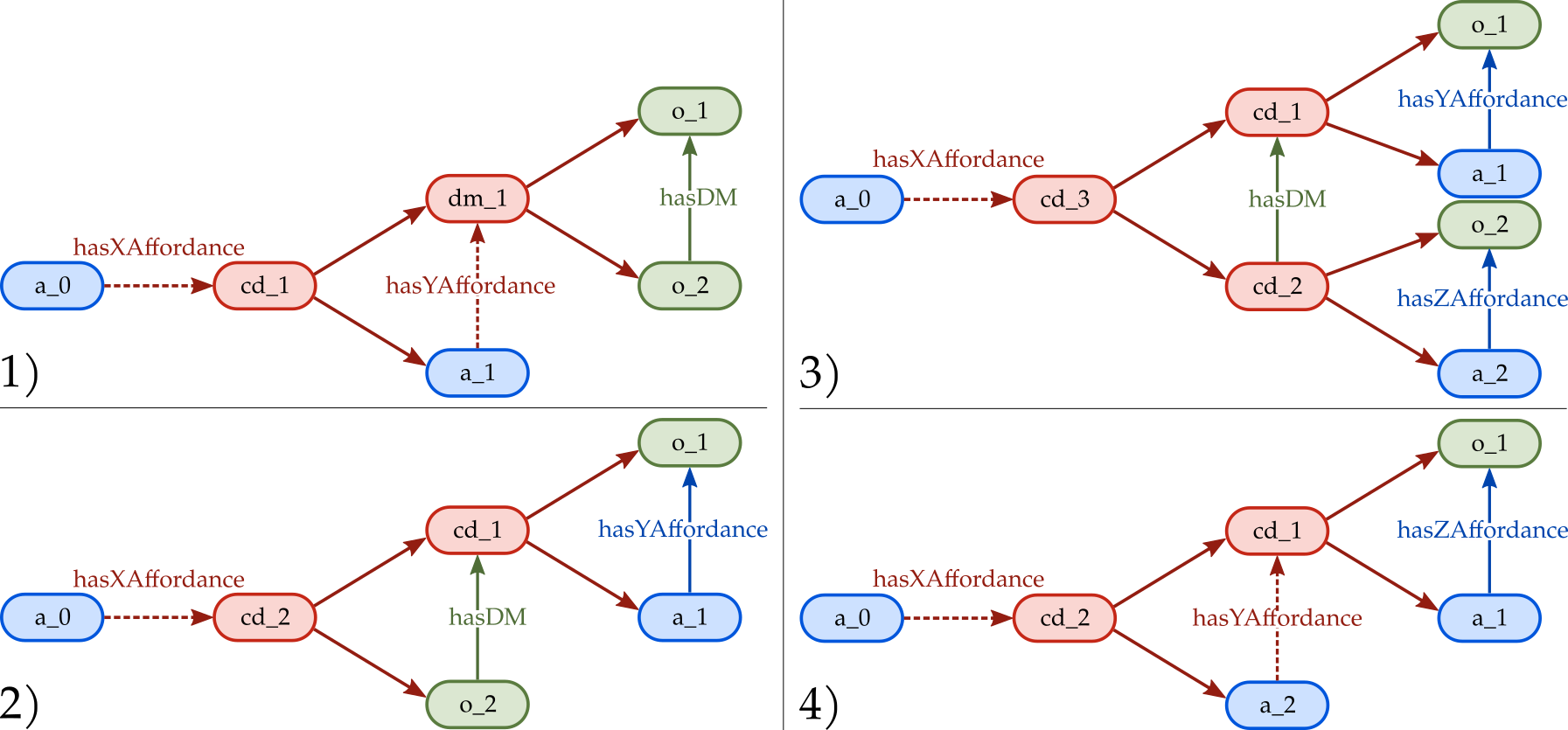}
    \caption{Four ontological patterns to represent: 1) cooperative affordance using a dispositional match, 2) cooperative affordance to collaborate, 3) cooperative affordance to coordinate, and 4) transitive cooperative affordance. The affordance properties X, Y, and Z are generic, thus can be different. Agents' capabilities and objects' dispositions are omitted for readability.}
    \label{fig:complex_patterns}
\end{figure*}

Using the elementary patterns and focusing on cooperative affordances, in this section we show how their combination can be used to represent complex situations, which lead to agent-exploitation. The second number of each subsection directly refers to the underlying representation depicted in Fig.~\ref{fig:complex_patterns}.

\subsection{Cooperative Affordance using a Dispositional Match}

Given the patterns introduced to represent an affordance from an agent toward a DM and a cooperative affordance in Fig.~\ref{fig:basic_patterns}.0b/0c, we can see that combinations of those patterns can arise. We therefore can represent that an agent is able to make use of another agent which in turn can actualize a combination of entities' dispositions. Indeed, one could consider an agent having an affordance toward a DM similarly to the single object case. This results in the creation of a CD (\textit{cd\_1}) representing the sum of the involved entities' dispositions (\textit{dm\_1}) and the interaction dispositions of the agent (\textit{a\_1}). An agent (\textit{a\_0}) capable of interacting with this latter can therefore have an affordance of actualizing the entities' dispositions through the other agent \textit{a\_1}. 

For instance, let us consider a scenario in which an agent needs to place a dirty piece of cutlery in the dishwasher to actualize the \textit{Washable} disposition. If this agent doesn't have the required capabilities to do so but another one holds them, then the cooperative affordance using the dispositional match between the dirty piece of cutlery and the dishwasher emerges.

\subsection{Cooperative Affordance to Collaborate}

The actualization of a disposition in a DM requires the agent to hold the required capabilities to interact with both the entities involved in the DM, thus the corresponding functional affordances toward each of those. However, situations in which several agents are present but do not own individually the required set of capabilities can occur. In order to actualize the disposition, the agents need to 
engage in a collaborative process since their collective holds the distributed capabilities (collaborative affordances).
Similarly to the aforementioned example, an agent (\textit{a\_1}) having an affordance toward an entity (\textit{o\_1}) creates a CD (\textit{cd\_1}), which embeds the entity's actualizable disposition. Thus, this compositional individual can be matched with another entity (\textit{o\_2}) if their dispositions are reciprocal, leading to the creation of a DM. However, as this match involves an agent and not only non-agentive entities, this newly created individual is a CD (\textit{cd\_2}). If an agent holds the required capabilities to interact with the other agent and to engage in the action on the other entity involved in the dispositional match (\textit{o\_2}), then a cooperative affordance emerges between this latter and the CD. This pattern enables to represent that \textit{a\_0} can collaborate with \textit{a\_1} to actualize the disposition of \textit{o\_1} via \textit{o\_2}, without directly acting upon all entities.

To illustrate, let us consider a situation in which one agent is capable of grasping the dirty piece of cutlery and the other one of using the dishwasher. In this context, one agent could communicate to the other one that they can use their respective affordances to actualize the \textit{Washable} disposition of the dirty cutlery with the \textit{Washing} disposition of the dishwasher. Given the reciprocity of the dispositions involved and each agent's functional affordance of acting upon the environment, a collaborative affordance emerges.

\subsection{Cooperative Affordance to Coordinate}

As we saw above, collaboration between agents can emerge from distributing the required affordances between several agents. For such a collaboration to occur, the agents need to interact with each other. However, another agent which is capable of interacting with both of the “acting” agents could also actualize the corresponding dispositions, without engaging in the action itself. This can be referred to as coordinating agents, and can be useful if acting agents cannot interact with one another directly, either for interaction modality compatibility, or for proximity/temporal reasons. Indeed, two agents (\textit{a\_1}, \textit{a\_2}) can respectively have a functional affordance toward entities (\textit{o\_1}, \textit{o\_2}). Those affordances respectively create a CD individual (\textit{cd\_1}, \textit{cd\_2}), representing the entity's dispositions and the agent's interaction disposition. If the entities' dispositions embedded in each CD are complementary, then they can be matched together in a DM-like manner. As the dispositional match occurs between two CD individuals, it results in the creation of another CD (\textit{cd\_3}) which represents both the agents' interaction dispositions and the involved entities' dispositions. Finally, an agent (\textit{a\_0}) having the capability to interact with both agents could then coordinate those agents to actualize the entities' dispositions by communicating its goal to each agent and how each of their involvement in the action can lead to the actualization of the dispositions.

To illustrate, let us consider the same scenario as in the aforementioned example, where each agent has a part of the required affordances to actualize the disposition of the dirty piece of cutlery. In this context, a third agent, which aims at washing the knife with the dishwasher could query each agent for an action (or communicate its goal) so that they both use their respective affordance. This results in a cooperative affordance to coordinate since the third agent doesn't engage directly in the action to actualize the desired disposition.

\subsection{Cooperative Affordance to Act by Transitivity}
“the ability to carry out a type of activity”
We have represented that an agent coordinating others can actualize an entity's disposition without engaging in the action, but one could envision that an agent could also do so in a transitive manner. Similarly to 0c, an agent (\textit{a\_1}) having an affordance toward an entity (\textit{o\_1}) creates a CD (\textit{cd\_1}) as they can be considered as a reciprocal match. Another agent (\textit{a\_2}) can actualize the disposition of the entity through interaction with the agent involved in the emergence of this CD.
Since this agent has an affordance toward the CD individual representing the composition of the actualizable entity and the interacting dispositions of the other agent, a new CD is created (\textit{cd\_2}). This latter encompasses the dispositions for interaction of the newly involved agent and can therefore be linked to an agent (\textit{a\_0}) capable of interacting with the intermediary agent (\textit{a\_2}) and not necessarily with the “acting” agent (\textit{a\_1}). Therefore, the actualization of the disposition occurs transitively with agents interacting with each other. The affordance relation generated illustrates that an agent can actualize an entity's dispositions via a chainlike process of interactions.

To illustrate, agents who are unable to interact directly with one another may require the involvement of a third party, acting as a translator. For instance, let us consider that the robot has the information that the dishwasher has been filled but cannot communicate with the agent that can start it. The presence of an intermediary agent which can interact with both therefore enables the robot to start the dishwasher without interacting with the “acting” agent.

\section{Combined Patterns: An Example}
\label{sec:combined}

\begin{figure}[!b]
    \centering
    \includegraphics[width=0.8\linewidth]{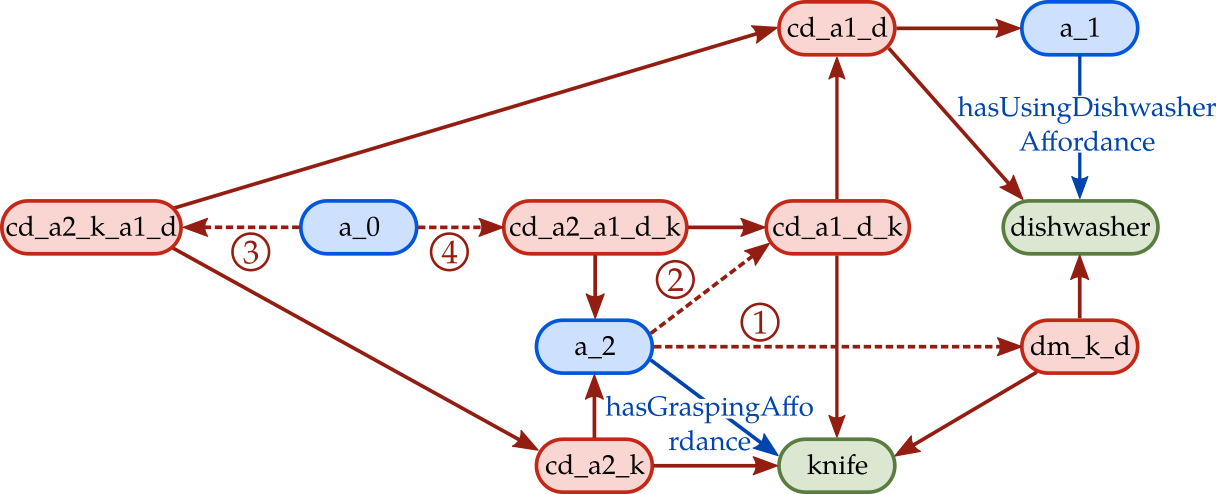}
    \caption{ 
    Affordance relations given several agents' points of view toward a similar actualization of a disposition. This pattern combines cooperative affordances using dispositional matches \textcircled{1}, to collaborate \textcircled{2}, to coordinate \textcircled{3} and to transitively act \textcircled{4}. Dotted lines represent the cooperative affordance relations. The dispositional match property, agents' capabilities, and object' dispositions have been omitted for readability purposes.}
    \label{fig:combined}
\end{figure}

In light of the elementary and complex patterns presented in the previous sections, one could grasp that such representations have the potential to provide a rich perspective of how agents might interact with each other to actualize the dispositions of entities. A comprehensive overview of the various pathways enabling such dispositional actualization is illustrated in Fig.~\ref{fig:combined} with the example of a dirty knife which requires washing from a dishwasher.

The goal of this example is to actualize the disposition of the knife of being \textit{Washable} given the reciprocal disposition \textit{Washing} of the dishwasher. The example illustrates how 3 agents in the environment can act together in order to wash the knife with the dishwasher. Actualizing the disposition given the pair of entities requires that agents have the capabilities to interact with both entities, hence grasping the knife and using the dishwasher. For this example, we will make the assumption that all agents can interact with each other and we chose to represent a subset of the affordance relations enabled by such a situation. However, a more exhaustive representation of this context would include more affordance relations.

Given the reciprocal dispositions of the dishwasher and the dirty knife, a new individual representing the sum of their complementary dispositions is created (\textit{dm\_k\_d}). If agent 2 has the required capabilities to grasp the knife and to use the dishwasher, an affordance relation can be created toward this dispositional match individual \textcircled{1}. Through this affordance relation, agent 2 can actualize the \textit{Washable} disposition of the knife with the dishwasher by itself, similarly to Fig.\ref{fig:complex_patterns}.1.

On the other hand, the fact that agent 1 has a functional affordance toward the dishwasher gives rise to a new CD individual (\textit{cd\_a1\_d}) which embeds the agent's interaction disposition and the actualizing disposition of the dishwasher. Since the knife has a reciprocal disposition to the dishwasher and that its disposition is embedded in the newly created CD individual, a dispositional match also occurs between this latter and itself (\textit{cd\_a1\_d\_k}). Thanks to the grasping and the interaction capabilities of agent 2, this agent can have the affordance to collaborate with agent 1\textcircled{2}, similarly to Fig.\ref{fig:complex_patterns}.2.

Moreover, since agents 1 and 2 have respectively the functional affordance of using the dishwasher and grasping the knife then it gives rise to new CD individuals embedding their interaction dispositions and respectively the knife and the dishwasher's dispositions (\textit{cd\_a2\_k} and \textit{cd\_a1\_d}). Given that those two individuals hold the reciprocal dispositions of the knife and the dishwasher, a dispositional match occurs between them. However, since this dispositional match happens between CD individuals (and not DMs), the resulting individual is a new CD individual (\textit{cd\_a2\_k\_a1\_d}). Considering that agent 0 can interact both with agents 1 and 2, it can therefore have the cooperative affordance of coordinating the two agents to wash the knife with the dishwasher \textcircled{3}, similarly to Fig.\ref{fig:complex_patterns}.3.

Lastly, we saw above that agent 1 has the affordance of using the dishwasher, agent 2 has the affordance to grasp the knife and that they can engage in a collaborative process initiated by agent 2 to actualize the knife's disposition \textcircled{2}. Therefore, from the affordance relation linking agent 2 to the CD individual representing the collaboration with agent 1 (\textit{cd\_a1\_d\_k}), a new CD individual emerges representing the interaction necessary to enable the collaboration between agent 1 and 2 (\textit{cd\_a2\_a1\_d\_k}). Therefore, through an interaction chain, agent 0 can query agent 2 to act with agent 1 in collaboration to wash the knife with the dishwasher. This results in a cooperative affordance for transitive action via a collaborative affordance \textcircled{4}, similarly to Fig.\ref{fig:combined}.4 and Fig.\ref{fig:complex_patterns}.2 combined.

As we saw above, multiple affordance relations can result in the actualization of the knife's disposition thanks to the dishwasher's reciprocal disposition from the point of view of the agents. The flexibility of the proposed patterns could allow to represent even more complex scenarios, in which each agent can have various ways of inducing a change in the environment given their objective.

\section{Conclusion and Future Work}
\label{sec:conclusion}

In this paper, we build upon the idea that affordances can emerge through the involvement of several agents in a given environment, such as a human-robot collaboration scenario. The action possibilities of agents are therefore not only restricted by the object's dispositions but also by the interaction between agents to bring a desired change to the environment, leading to the concept of \textbf{agent-exploitation}, which to the best of our knowledge has never been tackled. For this purpose, we \textbf{introduced and represented the concept of agent-exploitation through cooperative affordances}, building upon the similarity with the concept of dispositional match. The potency of this concept and its tractable representation has been highlighted thanks to \textbf{combinations of elementary patterns to represent complex cooperative situations}.

A future extension could be to refine the concept of collaborative affordances to support the notion of joint affordances. Indeed, while the current work allows to represent two agents that could act together with two objects given their complementary dispositions, it does not support the representation of two agents that could act jointly on the same object to collaborate. Such affordance could be for example to carry a heavy table together, where agents do not have the affordance individually toward this entity, but rather as a collective since this affordance stems from the combination of their capabilities. Another aspect which could be taken into consideration would be to represent by which means agents communicate with each other. This would enable an even more precise representation of the actual cooperative affordances that can occur.

While the current work focuses on the affordances schema, leaving aside the representation of agent capabilities and entities' dispositions on the examples, those representations have been built and are compatible with the proposed approach. Nevertheless, as we consider knowledge representation not only in a conceptual manner but also as a powerful tool for applicable purposes, we plan to leverage it for robotic applications to assess its effectiveness.

\begin{credits}
\subsubsection{\ackname} This work has been supported by the Effective Learning of Social Affordances (ELSA) project ANR-21-CE33-0019, and by the Artificial Intelligence for Human-Robot Interaction (AI4HRI) project ANR-20-IADJ-0006.

\subsubsection{\discintname}
\textbf{The authors have no competing interests to declare that are relevant to the content of this article.}
\end{credits}

\bibliographystyle{splncs04}
\bibliography{biblio}

\end{document}